Short Paper

Kawsar Noor[1,2], Katherine Smith[4], Julia Bennett[4], Jade O'Connell[4], Jessica Fisk[4], Monika Hunt[3], Gary Philippo[3], Teresa Xu[3], Simon Knight[3], Luis Romao[1,2], Richard JB Dobson[1,2,5,6,7*], Wai Keong Wong[2*]

*joint leadership

[1] Institute of Health Informatics, University College London, London, UK
[2] NIHR University College London Hospitals Biomedical Research Centre, University College London and University College London Hospitals NHS Foundation Trust, London, United Kingdom
[3] Business Intelligence Partners, University College London Hospitals NHS Foundation Trust, UK
[4] Gastrointestinal Services, University College London Hospitals NHS Foundation Trust, UK
[5] Health Data Research UK London, University College London, UK
[6] Department of Biostatistics and Health Informatics, Institute of Psychiatry, Psychology and Neuroscience, King's College London, UK
[7] NIHR Biomedical Research Centre, South London and Maudsley NHS Foundation Trust and King's College London, UK

# Predicting Clinical Intent from Free Text Electronic Health Records

## Abstract

**Background**
After a patient consultation, a clinician determines the steps in the management of the patient. A clinician may for example request to see the patient again or refer them to a specialist. Whilst most clinicians will record their intent as 'next steps' in the patient's clinical notes, in some cases the clinician may forget to indicate their intent as an order or request, e.g. failure to place the follow-up order. This consequently results in patients becoming 'lost-to-follow up' and may in some cases lead to adverse consequences.

**Objective**
Train a machine learning model to detect a clinician's intent to follow up with a patient from the patient's clinical notes.

**Methods**

Trained a machine learning model to detect clinical intent using a dataset of annotated clinical notes taken from the bariatric clinic from the University College London Hospitals (UCLH) NHS trust. A total of 3000 notes were annotated by three blinded annotators. This dataset was then used to train a natural language processing (NLP) multilabel classification model.

**Results**
Annotators systematically identified 22 possible types of clinical intent and annotated 3000 Bariatric clinical notes. The annotation process revealed a class imbalance in the labeled data and we found that there was only sufficient labeled data to train 11 out of the 22 intents. We used the data to train a BERT based multilabel classification model and reported the following average accuracy metrics for all intents: macro-precision: 0.91, macro-recall: 0.90, macro-f1: 0.90.

**Conclusions**
NLP models can be used to successfully detect clinical intent in clinical free text notes.

# Introduction

After a patient's consultation a clinician is responsible for planning and booking next steps. This might include requesting an imaging procedure,referring the patient to a specialist or even discharging the patient. All of these "intents" to follow up will typically be written in the patient's free text notes during the consultation by the clinician. The clinician/healthcare team is thereafter responsible for requesting/booking the item through the electronic healthcare (EHR) system. For example, if a clinician has referred the patient to a specialist there will also be a corresponding hospital order to make in order to complete the referral process. In some cases, however, clinicians fail to indicate their plans by formally placing such orders thus leading to the patient becoming 'lost in the system'.

Since in almost all circumstances, the intent has been recorded in the clinical notes we believe it is possible to build systems that can scan through these notes and detect the clinician's intent thereby enabling one to determine gaps between what a clinician has intended to do and what has been ordered/requested in the system.

In this paper we train an NLP based machine learning model to detect clinical intent in the patient's free text notes. We envisage that such models could be used as input for downstream systems capable of alerting clinicians of potentially lost-to-follow up patients, or even auto suggest clinical orders that might need requesting for a patient.

## Methods

### Overview

An extract of clinical notes was taken from the bariatrics clinic at UCLH. These notes included a wide variety of clinical notes that were generated for patients within the clinic between 2019-2021.

Analysts determined that there were a total of 22 different types of intents/actionable items that could be expressed within these notes. These included things such as requesting an imaging procedure, an outpatient appointment or even discharging the patient.

We randomly sampled 3000 documents from the data extract and asked three blinded annotators to annotate each document by highlighting phrases/sentences in a document and tagging each phrase with an intent. The annotated dataset was then used to train a multilabel text classification model.

### Dataset

The final annotated dataset revealed a class imbalance. There were a total of 2095 annotations with the mean number of annotations per intent of 99 (standard deviation 123). We chose to exclude all intents with less than 50 associated annotations. This resulted in dropping 11 out of the 22 intents. Our final dataset containing the 11 intents is captured in Figure 1.

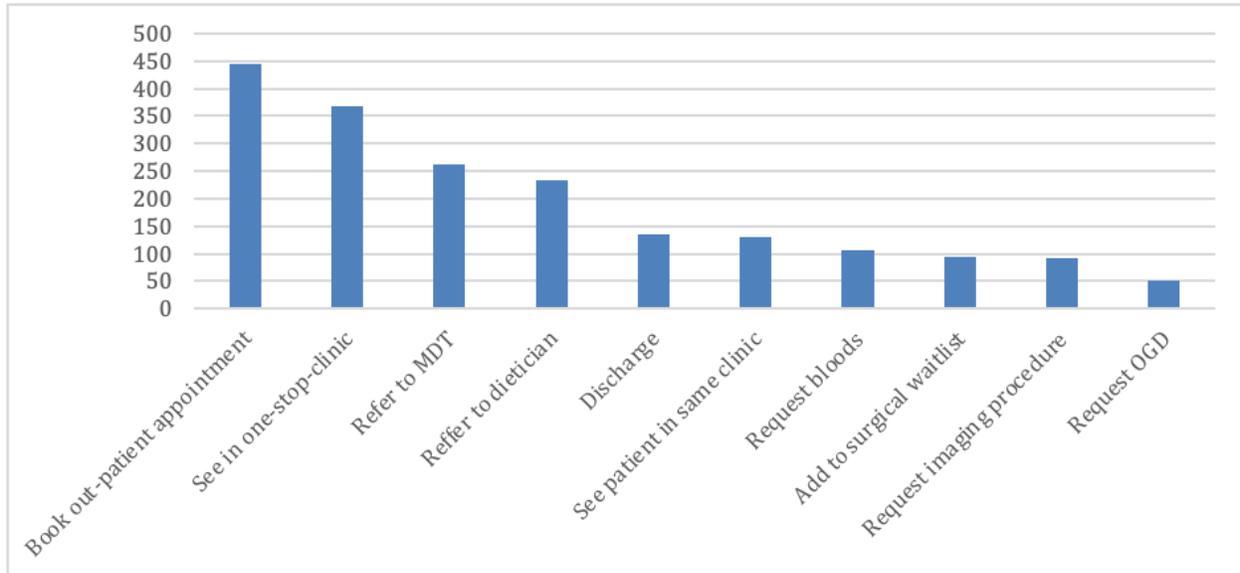

Figure1: No. of annotations per intent

The annotation process also revealed that intents were always expressed within a single sentence or phrase. This meant we could frame the problem as a sentence classification problem in which we are training a model to take an input sentence and predict one or more intents.

To facilitate this, we tokenized each document into sentences. We observed that the free text notes in our hospital records did not tokenize properly when using off-the-shelf tokenizers and so we extended the NLTK[1] sentence tokenizer to include special delimiting characters (e.g. bullet points) that we found were common markers of a new sentence within the records.

**Model**
In recent years NLP has made rapid progress due to the advances in transfer learning which is a technique where a deep learning model is trained on a large dataset (base model) and then fine-tuned to perform a similar task on a different dataset to produce a fine-tuned model. Transfer learning has been successfully used for many text-based bioinformatics problems to solve problems such as text classification, named entity recognition [2] as well as more recently text-generation problems [1].

One architecture that has been popular in this space is the BERT model [5] which is a language model that can be fine-tuned for numerous downstream NLP tasks. We use a model [4] that has been pretrained on open source deidentified EHR records from (MIMIC III corpus).

---

[1] https://www.nltk.org/

# Results

| Intent | Precision | Recall | F1 |
| --- | --- | --- | --- |
| Book out-patient appointment | 0.91 | 0.94 | 0.93 |
| Request imaging procedure | 0.94 | 0.89 | 0.91 |
| Request bloods | 0.89 | 0.85 | 0.87 |
| Request Oesophago-Gastro-Duodenoscopy | 0.87 | 0.84 | 0.85 |
| Add to surgical waitlist | 0.75 | 0.73 | 0.74 |
| See patient in same clinic | 0.957 | 0.92 | 0.94 |
| Discharge | 0.96 | 0.957 | 0.96 |
| Refer to dietician | 0.95 | 0.93 | 0.94 |
| Refer to multi-disciplinary team meeting | 0.95 | 0.97 | 0.96 |
| See in one-stop clinic | 0.99 | 0.99 | 0.99 |

Table1: Performance metrics per intent

We split our dataset into an 80:20 training:test split and conducted 5-fold validation. The results in Table 1 show the precision, recall and F1 scores for each intent. Precision is defined as (TP + TN) / (TP + FP), recall as TP/(TP+FP) and F1 as 2*precision*recall/ precision + recall. Where TP is true positives, TN true negatives, FP false positives and FN false negatives. We found that the macro-average precision was 0.91, recall 0.90, and F1 0.90.

# Discussion

In this paper we demonstrated that pretrained NLP models can be trained to detect a clinician's intention to follow up with a patient based on the patient's free text notes.

Whilst our model did not use any of the trust's data as part of the pretraining we expect that performance should increase if done so. This is something we would like to explore in future work as well as exploring alternative base model architectures.

# Acknowledgments

This study has been supported by the National Institute for Health Research University College London Hospitals Biomedical Research Centre, in particular by the NIHR UCLH/UCL BRC Clinical and Research Informatics Unit

Author Richard JB Dobson is supported by the following: (1) NIHR Biomedical Research Centre at South London and Maudsley NHS Foundation Trust and King's College London, London, UK; (2) Health Data Research UK, which is funded by the UK Medical Research Council, Engineering and Physical Sciences Research Council, Economic and Social Research Council, Department of Health and Social Care (England), Chief Scientist Office of the Scottish Government Health and Social Care Directorates, Health and Social Care Research and Development Division (Welsh Government), Public Health Agency (Northern Ireland), British Heart Foundation and Wellcome Trust; (3) The BigData@Heart Consortium, funded by the Innovative Medicines Initiative-2 Joint Undertaking under grant agreement No. 116074. This Joint Undertaking receives support from the European Union's Horizon 2020 research and innovation programme and EFPIA; it is chaired by DE Grobbee and SD Anker, partnering with 20 academic and industry partners and ESC; (4) the National Institute for Health Research University College London Hospitals Biomedical Research Centre; (5) the National Institute for Health Research (NIHR) Biomedical Research Centre at South London and Maudsley NHS Foundation Trust and King's College London; (6) the UK Research and Innovation London Medical Imaging & Artificial Intelligence Centre for Value Based Healthcare; (7) the National Institute for Health Research (NIHR) Applied Research Collaboration South London (NIHR ARC South London) at King's College Hospital NHS Foundation Trust.

# Conflicts of Interest

None declared

# References

[1] Qiao Jin, Bhuwan Dhingra, Zhengping Liu, William W. Cohen, and Xinghua Lu. 2019. Pubmedqa: A dataset for biomedical research question answering. In EMNLP.

[2] Kraljevic, Zeljko, et al. "Multi-domain clinical natural language processing with MedCAT: the medical concept annotation toolkit." Artificial Intelligence in Medicine 117 (2021): 102083.

[3] Jinhyuk Lee, Wonjin Yoon, Sungdong Kim, Donghyeon Kim, Sunkyu Kim, Chan Ho So, Jaewoo Kang, BioBERT: a pre-trained biomedical language representation model for biomedical text mining, *Bioinformatics*, Volume 36, Issue 4, 15 February 2020, Pages 1234–1240,

[4] Alsentzer E. et al. (2019) Publicly available clinical bert embeddings. In: *Proceedings of the 2nd Clinical Natural Language Processing Workshop, Minneapolis, MN, USA*. pp. 72–78. Association for Computational Linguistics.

[5] Devlin J. et al. (2019) Bert: pre-training of deep bidirectional transformers for language understanding. In: Proceedings of the 2019 Conference of the North American Chapter of the Association for Computational Linguistics: Human Language Technologies, Volume 1 (Long and Short Papers), Minneapolis, MN, USA. pp. 4171–4186. Association for Computational Linguistics.

## Abbreviations:

NLP: natural language processing
EHR: electronic healthcare records
UCLH: University College London Hospitals